\documentclass[conference]{IEEEtran}
\IEEEoverridecommandlockouts
\usepackage{xcolor}
\usepackage{cite}
\usepackage{amsmath,amssymb,amsfonts}
\usepackage{algorithmic}
\usepackage{graphicx}
\usepackage{textcomp}
\usepackage{xcolor}
\usepackage{multirow}
\def\BibTeX{{\rm B\kern-.05em{\sc i\kern-.025em b}\kern-.08em
    T\kern-.1667em\lower.7ex\hbox{E}\kern-.125emX}}

\usepackage{pstool}

\makeatletter
\let\MYcaption\@makecaption
\makeatother
\usepackage{subcaption}
\makeatletter
\let\@makecaption\MYcaption
\makeatother

\usepackage{soul}
\usepackage[outline]{contour}
\usepackage{booktabs}
\usepackage{tikz}
\usetikzlibrary{arrows.meta, arrows, patterns}
\usetikzlibrary{calc}
\usepackage{pgfplots}
\pgfplotsset{compat=newest}

\definecolor{tab:darkviolet}{RGB}{148, 0, 211}
\definecolor{tab:bggrey}{RGB}{163, 163, 163}
\definecolor{tab:k}{RGB}{0, 0, 0}
\definecolor{tab:hotpink}{RGB}{255, 105, 180}
\definecolor{tab:black}{RGB}{148, 0, 211}
\definecolor{tab:r}{RGB}{255, 0, 0}
\definecolor{tab:b}{RGB}{0, 0, 255}
\definecolor{tab:olivedrab}{RGB}{107, 142, 35}
\definecolor{tab:deepskyblue}{RGB}{0, 191, 255}
\definecolor{tab:lime}{RGB}{0, 255, 0}

\newcommand \colorindicator[1]{%
	\begingroup%
	\setul{0.25ex}{0.4ex}%
	\contourlength{0.2ex}%
	{\textcolor{#1}{\tiny{$\blacksquare \hspace{-.5mm} \blacksquare$}}}%
	\endgroup
}

\makeatletter
\def\ps@IEEEtitlepagestyle{%
  \def\@oddhead{\mycopyrightnotice}%
  \def\@oddfoot{}
  \def\@evenhead{\@IEEEheaderstyle\thepage\hfil\leftmark\hbox{}}\relax
  \def\@evenfoot{}%
}
\def\mycopyrightnotice{%
  \begin{minipage}{\textwidth}
  \centering \scriptsize
  \textcolor{red}{Copyright~\copyright~2022 IEEE. Personal use of this material is permitted. Permission from IEEE must be obtained for all other uses, in any current or future media, including reprinting/republishing this material for advertising or promotional purposes, creating new collective works, for resale or redistribution to servers or lists, or reuse of any copyrighted component of this work in other works. Accepted to be Published in: Proceedings of the 2022 Latin American Robotics Symposium (LARS), 2022 Brazilian Symposium on Robotics (SBR), and 2022 Workshop on Robotics in Education (WRE).}
  \end{minipage}
}
\makeatother
    
\begin{document}








\title{Dense Prediction Transformer for Scale Estimation in Monocular Visual Odometry\\

\thanks{This work was supported by the Coordination of Improvement of Higher Education Personnel -- CAPES (grant 88887.687888/2022-00).}
}

\author{\IEEEauthorblockN{1\textsuperscript{st} André O. Françani}
\IEEEauthorblockA{\textit{Autonomous Computational Systems Lab (LAB-SCA)}\\
\textit{Computer Science Division} \\
\textit{Aeronautics Institute of Technology}\\
São José dos Campos, SP, Brazil \\
andre.francani@ga.ita.br}
\and
\IEEEauthorblockN{2\textsuperscript{nd} Marcos R. O. A. Maximo}
\IEEEauthorblockA{\textit{Autonomous Computational Systems Lab (LAB-SCA)} \\ 
\textit{Computer Science Division} \\
\textit{Aeronautics Institute of Technology}\\
São José dos Campos, SP, Brazil \\
mmaximo@ita.br}
}


\maketitle

\let\thefootnote\relax\footnote{\\978-1-6654-6280-8/22/\$31.00
\textcopyright2022 IEEE}


\begin{abstract}
Monocular visual odometry consists of the estimation of the position of an agent through images of a single camera, and it is applied in autonomous vehicles, medical robots, and augmented reality. However, monocular systems suffer from the scale ambiguity problem due to the lack of depth information in 2D frames. This paper contributes by showing an application of the dense prediction transformer model for scale estimation in monocular visual odometry systems. Experimental results show that the scale drift problem of monocular systems can be reduced through the accurate estimation of the depth map by this model, achieving competitive state-of-the-art performance on a visual odometry benchmark. 

\end{abstract}

\begin{IEEEkeywords}
monocular visual odometry, scale estimation, deep learning, monocular depth estimation, vision transformer
\end{IEEEkeywords}

\section{Introduction}
\label{sec:intro}
Visual odometry (VO) is an accurate and classical process of estimating the camera pose and motion from a sequence of images. It is a tracking problem widely applied in mobile robots and autonomous vehicles \cite{scaramuzza2011VO}. 

Monocular visual odometry (MVO) systems use a single camera to capture the images, while the stereo VO algorithms utilize a stereo camera pair which allows the computation of feature depth between image frames. In this work, we deal with MVO mainly due to its simplicity, i.e. monocular systems have simpler hardware than stereo and are more accessible in society, especially via mobile devices. Nevertheless, MVO systems lack the depth information and scale of the objects since the three-dimensional (3D) objects in the world are projected into a two-dimensional (2D) image space. This implies that MVO algorithms can only estimate the relative motion, resulting in scale ambiguity. Furthermore, the scale errors during the motion estimation step are accumulated over time, leading to a scale drift, which is a critical factor in decreasing the accuracy of MVO methods \cite{scaramuzza2011VO}.

Recovering the scale information is challenging in mono\-cular systems and usually relies on a prior known absolute reference. The integration with other sensors, such as a global positioning system (GPS), wheel odometry, or an inertial measurement unit (IMU), can provide a reference scale \cite{scaramuzza2011VO}. Traditional solutions use local optimization, such as bundle adjustment (BA) \cite{triggs1999bundle} and loop closure (LC) \cite{mur2015orb} detection, to reduce the scale drift problem. Furthermore, researchers also use a ground plane estimation with a prior knowledge of the camera height \cite{zhou2019ground, tian2021accurate}, assumed to be stable during the motion, and the pixel depth map \cite{Yin_2017_ICCV} to estimate the scale and correct the computed translation, among other solutions \cite{botterill2012correcting, grater2015robust}. Deep learning techniques, e.g. transformer and convolutional neural networks (CNN), have shown high accuracy in estimating dense depth maps from single images \cite{godard2019digging, ranftl2021vision}.

In this paper, we will handle the scale recovery problem using the depth map estimated by a deep learning technique, more precisely by a transformer-based network. This work shows that the scale estimated by a transformer-based network yields competitive performance compared to the state-of-the-art solutions on the KITTI odometry benchmark \cite{Geiger2012CVPR}.

The remaining of this paper is organized as follows. Section~\ref{sec:dpt} introduces the transformer-based network used to predict the monocular depth map. 
Section~\ref{sec:mvo} explains the monocular visual odometry components and presents our pipeline model.
Section~\ref{sec:eval} details the KITTI benchmark dataset and the metrics used to evaluate and compare our model with other state-of-the-art algorithms.
Section~\ref{sec:results} shows the experimental results on the VO benchmark. 
Section~\ref{sec:discussion} discusses the achieved results.
Finally, Section~\ref{sec:conclusion} concludes and shares our ideas for future work.

\section{Depth estimation through dense prediction transformer}
\label{sec:dpt}

Transformer \cite{vaswani2017attention} is a state-of-the-art architecture based on attention mechanisms that outperformed recurrent neural networks (RNN) and long short-term memory cells (LSTM) in natural language processing (NLP) tasks. Since then, transformer-based architectures have begun
to emerge, like BERT \cite{devlin2018bert} and GPT \cite{radford2018improving}. After the success of transformer-based architectures in NLP tasks, researchers started to apply transformer-based networks to vision problems. Dosovitskiy \emph{et al.} \cite{dosovitskiy2020image} presented the vision transformer (ViT) model, reaching state-of-the-art performance on several image recognition benchmarks \cite{dosovitskiy2020image}. 

Regarding the monocular depth estimation from a single 2D image, Ranftl \emph{et al.} \cite{ranftl2021vision} introduced the dense prediction transformer (DPT) model, reporting an improvement in relative performance compared to the CNN-based state-of-the-art \cite{ranftl2021vision}. For this reason, we chose the DPT network to estimate the depth map since an accurate depth estimation contributes to a better scale estimation \cite{yin2017scale}.

\begin{figure}[!tbp]
\centering
    \begin{subfigure}[]{0.24\textwidth}
         \includegraphics[width=\textwidth]{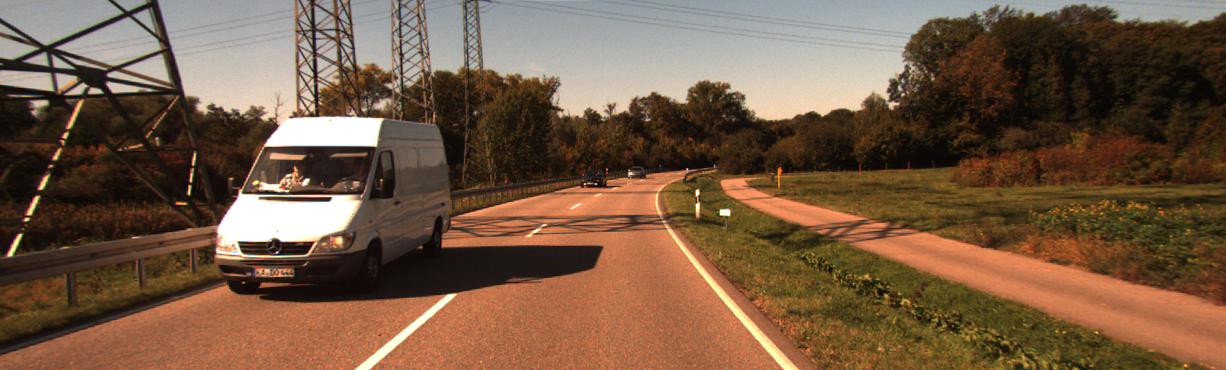}
    \end{subfigure}
    \begin{subfigure}[]{0.24\textwidth}
         \includegraphics[width=\textwidth]{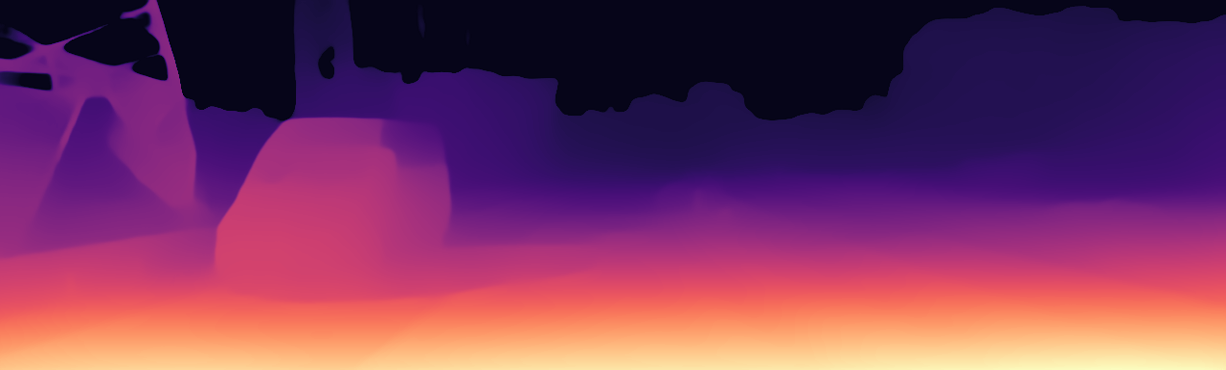}
    \end{subfigure}
    \par\smallskip
    \begin{subfigure}[]{0.24\textwidth}
         \includegraphics[width=\textwidth]{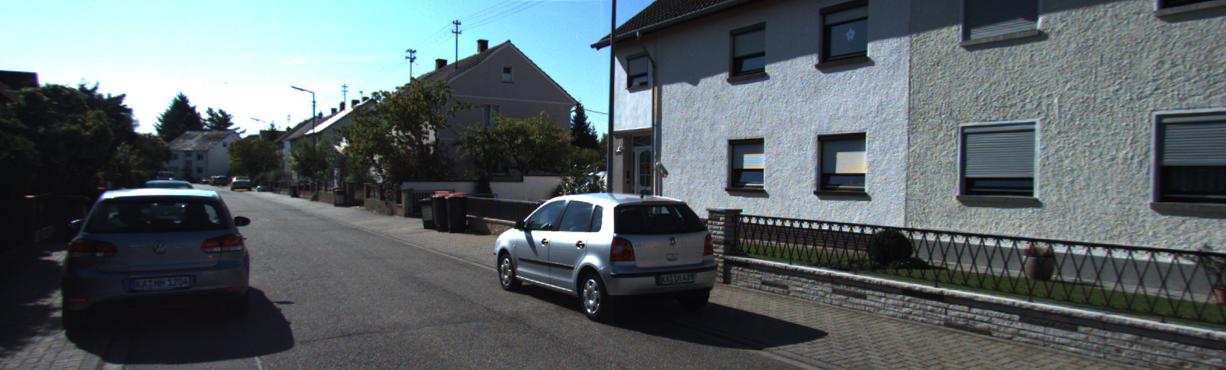}
    \end{subfigure}
    \begin{subfigure}[]{0.24\textwidth}
         \includegraphics[width=\textwidth]{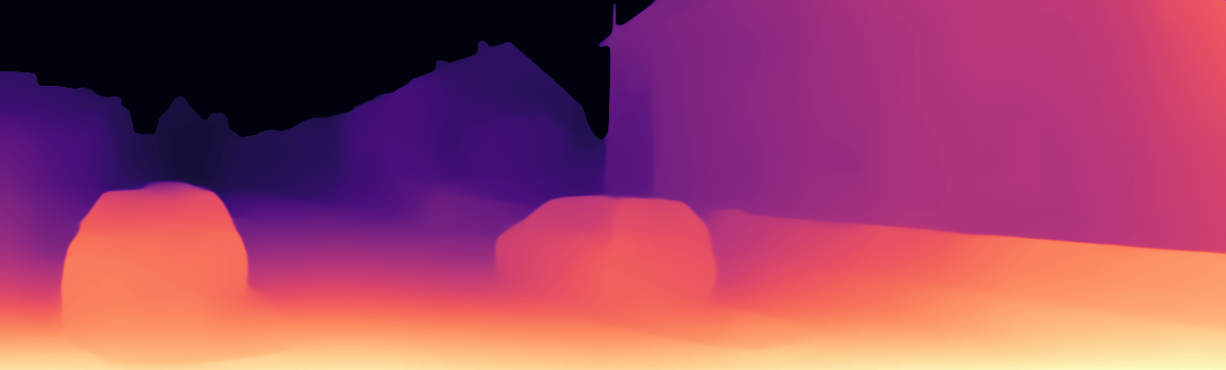}
    \end{subfigure}
    \caption{Original KITTI RGB images (left) and monocular depth estimated by DPT (right). Dark blue colors denote further objects from the camera, while light red/yellow colors indicate closer objects.}
    \label{fig:depth}
\end{figure}

The DPT model utilizes a ViT as the backbone. The input image is divided into image patches, which are embedded as flattened representations of features extracted from a ResNet 50 network \cite{ranftl2021vision}. The embedding is augmented with a special positional token called \emph{readout} token \cite{ranftl2021vision}. The embedding process with a CNN feature extractor makes the model hybrid (DPT-Hybrid architecture in \cite{ranftl2021vision}). The image patches are equivalent to ``words'' in NLP tasks, and the embedded patches will be referred to as tokens, following the original nomenclature of the transformer architecture. The set of tokens passes through a sequence of transform layers composed of multi-headed self-attention blocks \cite{vaswani2017attention}. The output tokens of the transformer layers are reassembled by a Reassemble operation \cite{ranftl2021vision}, followed by fusion blocks \cite{ranftl2021vision} that combine the features progressively. The DPT-Hybrid architecture uses the features extracted from layers 9 and 12 of the ResNet 50 in the embedding step. The interested reader is referred to the original paper for a more detailed explanation \cite{ranftl2021vision}. From now on, the DPT-Hybrid model will be referred to only as DPT in this paper. Figure~\ref{fig:depth} shows the depth map estimated by the DPT model.

\section{Monocular visual odometry}
\label{sec:mvo}

Monocular visual odometry aims to estimate the agent's motion incrementally using the images from a single camera, pose after pose \cite{scaramuzza2011VO}. The motion $\mathbf{T}_k \in \mathbb{R}^{4\times4}$ is defined as 
$$\mathbf{T}_k = \begin{bmatrix}
                \mathbf{R}_k & \mathbf{t}_k  \\
                \mathbf{0}   & \mathbf{1}  \\
                \end{bmatrix},  $$
where $k$ is the current time instant, $\mathbf{R}_k \in SO(3)$ is the rotation matrix which describes the camera rotation from instant $k-1$ to $k$, and $\mathbf{t}_k \in \mathbb{R}^{3 \times 1}$ is the translation vector that depicts the translation from instant $k-1$ to $k$. Furthermore, the projection properties of the camera should also be considered, given by the intrinsic calibration matrix
$$\mathbf{K} = \begin{bmatrix}
                f_x & 0   & c_x \\
                0   & f_y & c_y \\
                0   & 0   & 1
                \end{bmatrix},$$
where $(f_x, f_y)$ is the focal length, and $(c_x, c_y)$ is the principal point (optical center) \cite{solem2012programming}.

Before estimating the motion $\mathbf{T}_k$, features must be detected and matched over the image frames. This step is defined in Subsection~\ref{subsec:feat}. In Subsections~\ref{subsec:E_matrix} and \ref{subsec:pnp}, two methods for estimating motion are detailed, one by decomposing the essential matrix and the other by minimizing the reprojection error.

\subsection{Feature detection and matching}
\label{subsec:feat}
The feature detection step consists of detecting interest points in the images, such as the corners. These points are called keypoints or features, and they should be easily detectable in the next frames, so the feature matching can be applied.

The Features from Accelerated Segment Test (FAST) \cite{rosten2006machine} is a corner detection algorithm with low computational cost, allowing real-time applications. The corners are tested by the number of contrasting pixels in a circle around a potential corner center, and detected as corners with the aid of a decision tree classifier. The contrast is a comparison of the pixel values around the center, checking if they are brighter or lighter, given a threshold. Since many potential corners are identified, a non-maximal suppression is applied to preserve only the most suitable corner candidates \cite{rosten2006machine}.

Following the detection step, the keypoints in a frame are compared with all other keypoints in the next frame using a similarity measurement \cite{scaramuzza2011VO}. The sparse features set are matched by calculating the optical flow between the frames using the iterative Lucas-Kanade method \cite{bouguet2001pyramidal}. 

\begin{figure}[tbp]
\centering
    \begin{subfigure}[]{0.48\textwidth}
         \includegraphics[width=\textwidth]{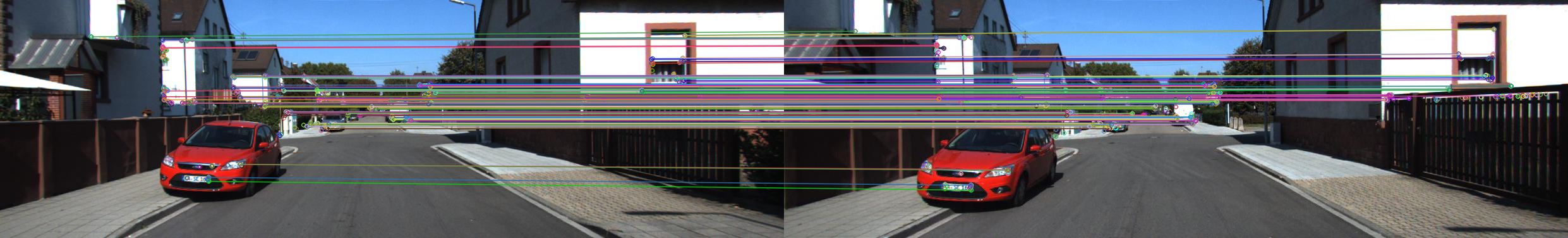}
    \end{subfigure}
    \par\smallskip
    \begin{subfigure}[]{0.48\textwidth}
         \includegraphics[width=\textwidth]{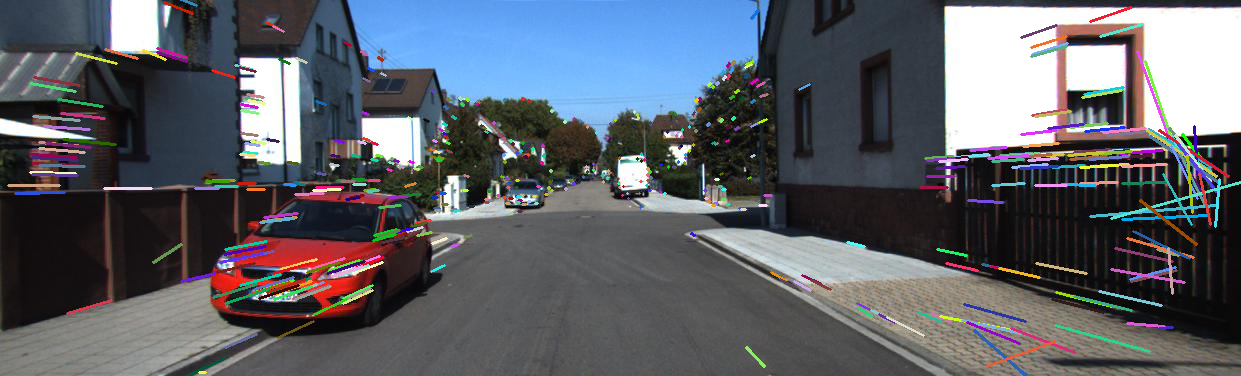}
    \end{subfigure}
    \par\smallskip
    \caption{Keypoint detection and matching between two consecutive frames of KITTI dataset. The upper figure shows the matched features between the frames side by side, while the lower depicts the keypoints tracked from the first to the second frame.}
    \label{fig:matching}
\end{figure}

\subsection{Motion estimation via essential matrix}
\label{subsec:E_matrix}

Motion estimation is the core component of VO algorithms. A classical geometry-based algorithm employs epipolar constraints for solving the essential matrix $\mathbf{E}_k$ or the fundamental matrix $\mathbf{F}_k$ at instant $k$, which are related by the intrinsic calibration matrix $\mathbf{K}$:
\begin{equation}
    \mathbf{F}_k = \mathbf{K^{-T}} \mathbf{E}_k \mathbf{K^{-1}}.
\end{equation}

The essential matrix $\mathbf{E}_k$ is related to the camera motion by the following:
\begin{equation}
\label{eq:Ematrix}
    \mathbf{E}_k \simeq [\mathbf{t}_k]_{\times} \mathbf{R}_k, 
\end{equation}
where 
$$[\mathbf{t}_k]_{\times} = \begin{bmatrix}
                0    & -t_z & t_y  \\
                t_z  & 0    & -t_x \\
                -t_y & t_x  & 0
                \end{bmatrix}$$
is the skew symmetric matrix of translation vector $\mathbf{t}_k = [t_x, t_y, t_z]$, and $\simeq$ indicates the estimation up to a scale factor \cite{scaramuzza2011VO}.

Let $\mathbf{I}_{k-1}$ and $\mathbf{I}_k$ be two consecutive image frames from a calibrated camera. Given the 2D normalized matched features coordinates $\mathbf{x}_k$ of image $\mathbf{I}_k$, and $\mathbf{x}_{k-1}$ of image $\mathbf{I}_{k-1}$, the epipolar constraint leads to the following equation:
\begin{equation}
    \mathbf{x}_{k}^T \mathbf{E}_k \mathbf{x}_{k-1} = 0.
\end{equation}
Then, the camera motion $\mathbf{T}_k$ can be estimated from 2D-to-2D feature correspondences by decomposing the essential matrix \cite{scaramuzza2011VO, nister2004efficient}. The decomposition leads to four possible solutions, but the correct solution can be estimated by checking the \emph{chirality} condition, i.e. if the triangulation of the feature points are in front of both cameras \cite{scaramuzza2011VO, zhan2021df}. 

The estimation of the essential matrix using epipolar constraints may suffer motion degeneracy when the camera's translation is small, or the motion is a pure rotation, and structure degeneracy when the viewed scene structure is planar \cite{torr1999problem}. In these cases, the estimation is unstable and the epipolar constraint is unsuitable. Furthermore, the translation recovered is up to an unknown scale factor, as seen in (\ref{eq:Ematrix}).

\subsubsection{Scale estimation}

There are several ways to estimate the relative scale in MVO and correct the translation vector. Popular methods use the prior knowledge of the camera height, and a road plane modeling \cite{zhou2019ground}. The depth information can also be used to estimate the scale. Zhan \emph{et al.} \cite{zhan2021df} proposed a simple scale recovery algorithm based on the depth estimated by a CNN model, aligning the triangulated depth with the deep learning estimated depth map. In this case, the deep learning model is assumed to be the ``depth sensor'', so the true 3D structures are assumed to be known. Let $D^{\prime}_i$ be the depth estimated by triangulation, and $D_i$ be the depth estimated by the deep learning model, where $i$ is the $i$-th matched keypoint, $i = 0, 1, \cdots, N$, and $N$ is the number of matched keypoints after prior outlier removal from the triangulation step. The scale is obtained by a random sample consensus (RANSAC) \cite{fischler1981random} regressor with the depth ratio vector $\mathbf{D} = \left[\frac{D^{\prime}_1}{D_1}, \frac{D^{\prime}_2}{D_2}, \cdots, \frac{D^{\prime}_N}{D_N} \right]^T$ as input, so the depths are aligned.

\subsection{Motion estimation via PnP}
\label{subsec:pnp}

Perspective from $n$ points (PnP) is an alternative method to the epipolar geometry when the 3D-to-2D correspondences are given. With the depth map estimated by the deep learning model, the depth of the keypoints are assumed to be known. Therefore, knowing the 3D point $\mathbf{P}_{k-1}$ and its 2D projection $\mathbf{x}_k$ in the next instant, the PnP estimates the camera motion $\mathbf{T}_k$ by minimizing the reprojection error \cite{scaramuzza2011VO}:
\begin{equation}
    \mathbf{T}_k = \arg \min_{\mathbf{T}_k} \sum _i  \left \| \hat{\mathbf{x}}_{k-1}^{i} - \mathbf{x}_k^{i} \right \|_{2}, 
\end{equation}
where the superscript $i$ indicates the $i$-th feature, $$\hat{\mathbf{x}}_{k-1}^{i} =  \mathbf{K}\left(\mathbf{R} \mathbf{P}_{k-1}^{i} + \mathbf{t}\right)$$ is the reprojection of the 3D point $\mathbf{P}_{k-1}^{i}$ in image $\mathbf{I}_{k}$, and $\left \| \cdot \right \|{_2}$ is the $\text{L}^2\text{-norm}$.


\subsection{Method selection}

As discussed in Subsection~\ref{subsec:E_matrix}, there are cases where decomposing the essential matrix to estimate motion is not the most appropriate method, mainly when motion and structure degeneracy occur. The geometric robust information criterion (GRIC) \cite{torr1999problem} is a statistical model that can identify such cases where degeneracy happens. Torr \emph{et al.} \cite{torr1999problem} computed the GRIC for the fundamental matrix $\mathbf{F}$ $(\text{GRIC}_\text{F})$ and the homography matrix $\mathbf{H}$ ($\text{GRIC}_{\text{H}}$). The smaller the GRIC score, the better the method explains the data. 
The GRIC function is defined as follows \cite{torr1999problem}:
\begin{equation}
    \text{GRIC} = \sum \rho\left(e_i^{2}\right) + \log(4) d n + \log(4n) k,
\end{equation}
where
$$
\rho\left(e_i^{2}\right) = \min \left(\frac{e^2}{\sigma^2}, 2(r-d) \right)
$$
is a robust function of the residuals $e_i$, $d$ is the dimension of the structure ($d=2$ for $\mathbf{H}$, $d=3$ for $\mathbf{F}$), $n$ is the number of matched features, $k$ is the number of motion model parameters ($k=7$ for $\mathbf{F}$, $k=8$ for $\mathbf{H}$), $r$ is the data dimension (4 for two views), and $\sigma$ is the standard deviation of the measurement error.

In this work, we follow \cite{zhan2021df} and apply the PnP method instead of the Homography method when $\text{GRIC}_\text{F} > \text{GRIC}_{\text{H}}$, since we have the depth information from the DPT model. Therefore, GRIC will behave as a method selector, switching to PnP when the essential matrix is detected as unsuitable to estimate the camera motion.

\subsection{DPT-VO pipeline}

\begin{figure}[tbp]
    \centering
    \includegraphics[width=0.49\textwidth]{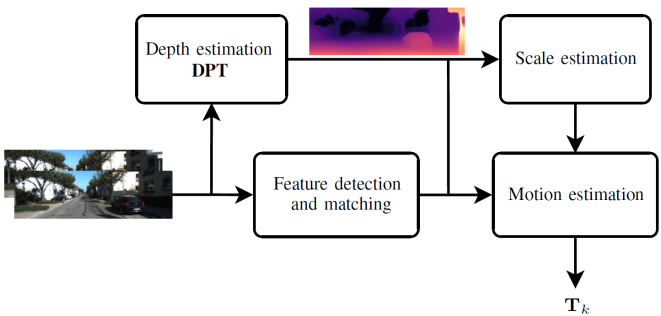}
    \caption{DPT-VO pipeline. A pair of image frames (at instants $k$ and $k-1$) are used as input to the feature detector, while only a single frame is input to the monocular depth estimator, performed by a DPT model. The features are matched between the frames and used for scale recovery and motion estimation.}
    \label{fig:pipeline}
\end{figure}

The pipeline of our MVO algorithm is called DPT-VO since we use a DPT model to estimate the depth used to recover the scale in the MVO task. Figure~\ref{fig:pipeline} shows a schematic diagram of the DPT-VO pipeline, which contain the main block components of VO system, namely: feature detection and matching block, depth and scale estimation blocks, and motion estimation block.

It is worth mentioning that the motion estimation block comprises the GRIC to select between the essential matrix and PnP methods to estimate the pose. Furthermore, the estimated scale through the depth map only corrects the motion when the essential matrix method is applied.


\section{Evaluation}
\label{sec:eval}

This section presents more details on the dataset and the metrics used to evaluate the MVO algorithms.

\subsection{KITTI dataset}

The KITTI odometry dataset \cite{Geiger2012CVPR} is a benchmark for developing and evaluating VO algorithms. A stereo camera mounted on a vehicle moving on the road captures the images during the ride. Although stereo images are available, we use only the images acquired from the left camera to implement our monocular system. The dataset comprises 22 sequences, from which 11 are provided as the trajectories with their ground truths, i.e., the camera position and orientation for each frame. The images are sampled and recorded at 10 Hz, and a GPS provides the ground truth. The remaining 11 sequences have no ground truth available to the community.

The trajectories vary from city, residential, and road scenes, with pedestrians, moving and parked cars, bicycles, etc. Moreover, the length of the trajectories in the sequences are not the same, and the car speed can range from 0 to 90 km/h. This makes visual odometry challenging, mainly regarding feature detection in high-speed and steep curve situations.

\subsection{Evaluation metrics}

The metrics are fundamental to evaluate methods and to compare the results with other approaches. For this reason, we evaluate the MVO performance using the metrics reported in KITTI odometry listed below: 
\begin{itemize}
    \item $t_{err}$: average translational error, measured in percentage (\%);
    \item $r_{err}$: average rotational error, measured in degrees per 100 meters (º/100 m);
    \item ATE: absolute trajectory error, given in meters by the root mean squared error (RMSE) between the estimated camera pose and the ground truth;
    \item RPE: relative pose error for rotation and translation, measured frame-to-frame and given in degrees (º) for the rotation and in meters (m) for the translation.
\end{itemize}
The average translational and rotational errors are computed for all possible subsequences of length $(100, 200, \cdots, 800)$ meters. 

It is worth noting that prior works applied certain degrees of alignment in the evaluation step since monocular methods suffer from scale ambiguity to match the real-world scale. Thus, an optimization transformation might be applied to align the predictions to the ground truth poses, such as 7DoF and 6DoF optimization.



\section{Experimental results}
\label{sec:results}

\begin{figure}[!tbp]
    \begin{subfigure}[]{0.234\textwidth}
         \includegraphics[width=0.99\textwidth]{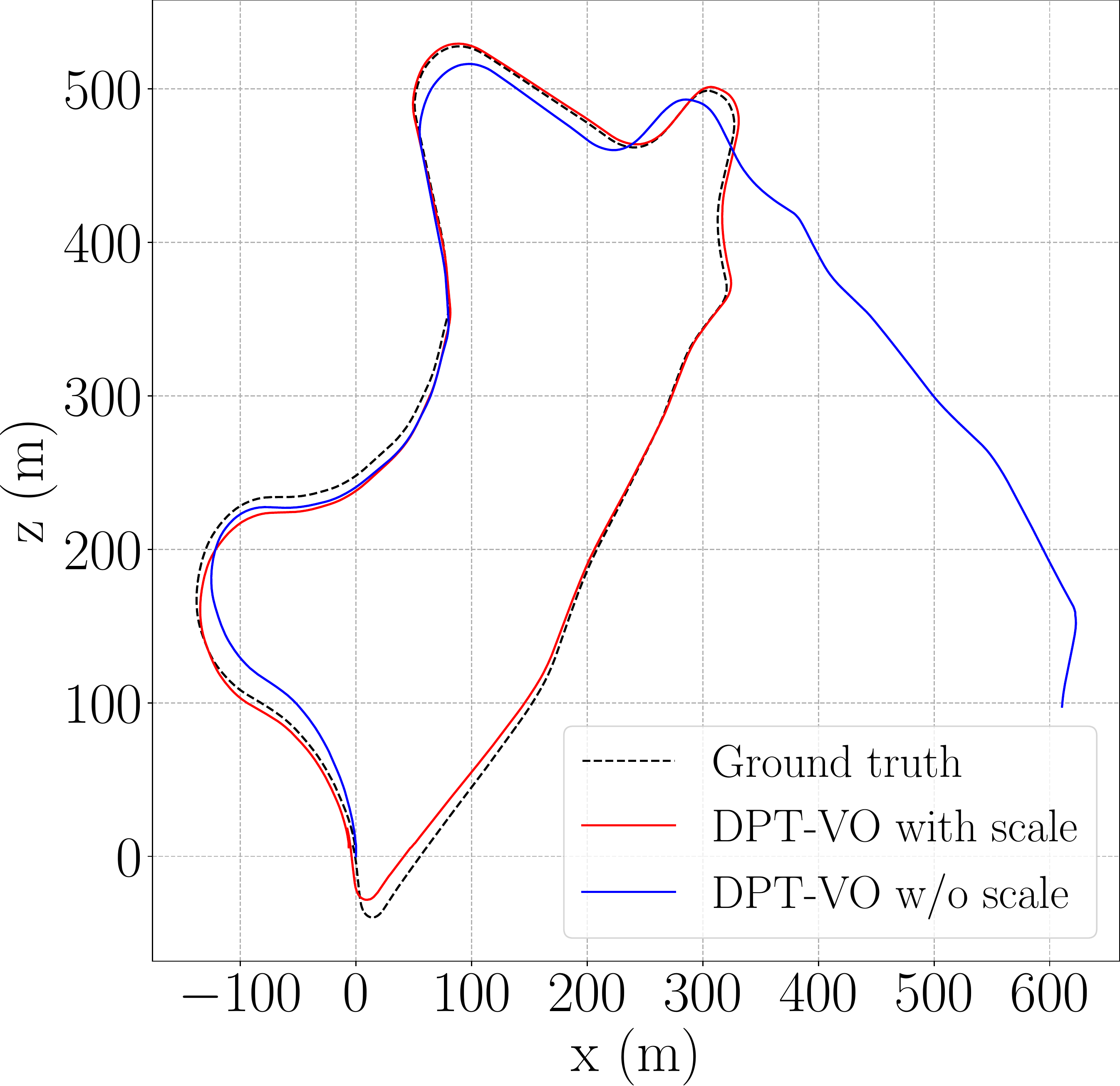}
         \caption{Sequence 09}
    \end{subfigure}
    \begin{subfigure}[]{0.234\textwidth}
         \includegraphics[width=\textwidth]{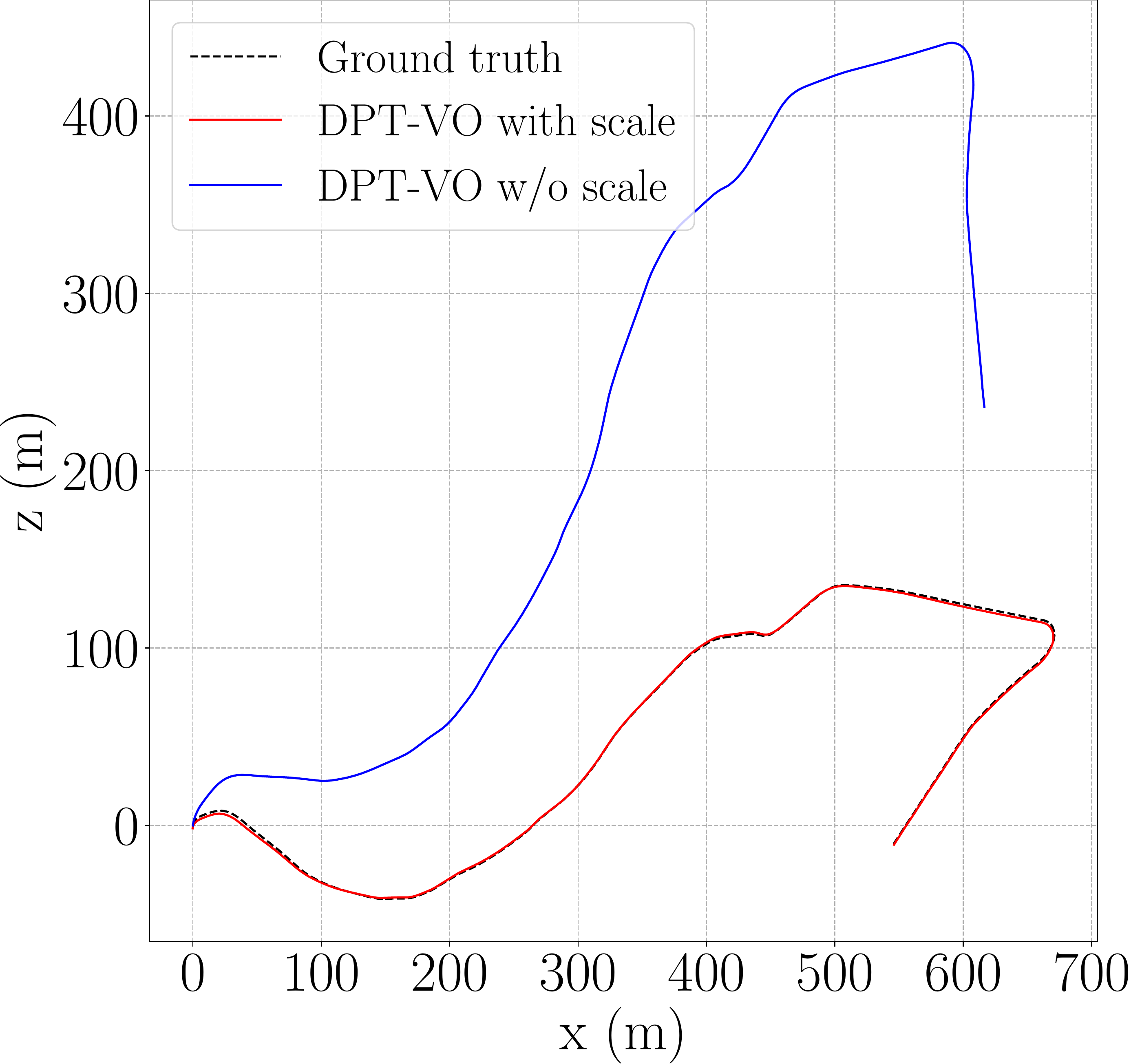}
         \caption{Sequence 10}
    \end{subfigure}
    \caption{Sequences 09 (a), and 10 (b) of the KITTI odometry dataset using the DPT-VO with and without the scale estimation obtained from the depth map.}
    \label{fig:plot_scale}
\end{figure}

To evaluate the applicability of the DPT network in estimating scale in visual odometry tasks, a qualitative experiment was carried out, shown in Figure~\ref{fig:plot_scale}, by applying our MVO pipeline to four KITTI sequences with and without the scale estimated from the depth of DPT, and without 7DoF optimization. In this example, the scale is set to the default value of 1.0 for the case without scale, so the translation vectors are not changed. The scaled trajectory is depicted in red, while the blue line represents the case without the scale estimation step, both compared to the ground truth in dashed black.
The DPT model used in this experiment was fine-tuned on KITTI, using the weights available in the original repository\footnote{https://github.com/isl-org/DPT}. 
As a result, it can be seen in Figure~\ref{fig:plot_scale} that scale estimation is essential to address the scale drift in MVO systems. In addition, the depth map obtained by DPT was accurate enough and useful to estimate the scale.

Next, a quantitative experiment was conducted comparing our DPT-VO with state-of-the-art models in MVO, such as ORB-SLAM2 without LC and DF-VO trained on stereo images. The ORB-SLAM2 algorithm might depend on the initialization, so the reported results come from the least ATE obtained in 3 runs.
Finally, the metrics were computed by the Python KITTI evaluation toolbox\footnote{https://github.com/Huangying-Zhan/kitti-odom-eval} in all 11 KITTI sequences with ground truth. For a fair comparison, the proper alignment optimizations were applied (7DoF and 6DoF), as typically done in the literature \cite{mur2015orb, zhan2021df}, and both the ORB-SLAM2 without LC and DF-VO results were taken from the DF-VO repository\footnote{https://github.com/Huangying-Zhan/DF-VO}. The quantitative results are presented in Table~\ref{tab:metrics}, and the best values of each metric for each sequence are highlighted in bold.
\begin{table}[tbp]
\caption{Metrics obtained by the selected VO methods for 11 KITTI sequences with ground truth. The best values are highlighted in bold.}
\label{tab:metrics}
\begin{tabular}{c|llllll}
\hline\hline
\textbf{Seq.} &
  \multicolumn{1}{c}{\textbf{Method}} &
  \multicolumn{1}{c}{\textbf{\begin{tabular}[c]{@{}c@{}}$t_{err}$ \\ (\%)\end{tabular}}} &
  \multicolumn{1}{c}{\textbf{\begin{tabular}[c]{@{}c@{}}$r_{err}$ \\ (º/100m)\end{tabular}}} &
  \multicolumn{1}{c}{\textbf{\begin{tabular}[c]{@{}c@{}}ATE\\ (m)\end{tabular}}} &
  \multicolumn{1}{c}{\textbf{\begin{tabular}[c]{@{}c@{}}RPE\\ (m)\end{tabular}}} &
  \multicolumn{1}{c}{\textbf{\begin{tabular}[c]{@{}c@{}}RPE\\ (º)\end{tabular}}} \\ \hline\hline
\multirow{3}{*}{00} & ORB-SLAM2 & 11.43          & 0.58          & 40.65           & 0.169          & 0.079          \\
                    & DPT-VO    & \textbf{1.44}  & \textbf{0.40} & \textbf{8.68}   & 0.038          & 0.059          \\
                    & DF-VO     & 2.01           & 0.61          & 12.17           & \textbf{0.025} & \textbf{0.055} \\ \hline
\multirow{3}{*}{01} & ORB-SLAM2 & 107.57         & \textbf{0.89} & 502.20          & 2.970          & 0.098          \\
                    & DPT-VO    & \textbf{43.44} & 1.81          & \textbf{123.67} & 1.745          & 0.120          \\
                    & DF-VO     & 40.02          & 0.47          & 342.71          & \textbf{0.854} & \textbf{0.052} \\ \hline
\multirow{3}{*}{02} & ORB-SLAM2 & 10.34          & \textbf{0.26} & 47.82           & 0.172          & 0.072          \\
                    & DPT-VO    & 2.47           & 0.48          & \textbf{14.80}  & 0.055          & 0.050          \\
                    & DF-VO     & \textbf{2.32}  & 0.48          & 17.59           & \textbf{0.030} & \textbf{0.045} \\ \hline
\multirow{3}{*}{03} & ORB-SLAM2 & \textbf{0.97}  & \textbf{0.19} & \textbf{0.94}   & 0.031          & 0.055          \\
                    & DPT-VO    & 6.471          & 1.40          & 8.22            & 0.138          & 0.067          \\
                    & DF-VO     & 2.22           & 0.30          & 1.96            & \textbf{0.021} & \textbf{0.038} \\ \hline
\multirow{3}{*}{04} & ORB-SLAM2 & 1.30           & 0.27          & 1.27            & 0.078          & 0.079          \\
                    & DPT-VO    & 5.09           & 0.37          & 5.35            & 0.196          & 0.035          \\
                    & DF-VO     & \textbf{0.74}  & \textbf{0.25} & \textbf{0.70}   & \textbf{0.026} & \textbf{0.029} \\ \hline
\multirow{3}{*}{05} & ORB-SLAM2 & 9.04           & \textbf{0.26} & 29.95           & 0.140          & 0.058          \\
                    & DPT-VO    & \textbf{1.30}  & 0.35          & \textbf{3.30}   & 0.034          & 0.039          \\
                    & DF-VO     & \textbf{1.30}  & 0.30          & 4.94            & \textbf{0.018} & \textbf{0.035} \\ \hline
\multirow{3}{*}{06} & ORB-SLAM2 & 14.56          & \textbf{0.26} & 40.82           & 0.237          & 0.055          \\
                    & DPT-VO    & \textbf{0.81}  & \textbf{0.26} & \textbf{1.87}   & 0.036          & 0.032          \\
                    & DF-VO     & 1.42           & 0.32          & 3.73            & \textbf{0.025} & \textbf{0.030} \\ \hline
\multirow{3}{*}{07} & ORB-SLAM2 & 9.77           & 0.36          & 16.04           & 0.105          & 0.047          \\
                    & DPT-VO    & 2.93           & 1.02          & 5.98            & 0.038          & 0.042          \\
                    & DF-VO     & \textbf{0.72}  & \textbf{0.35} & \textbf{1.06}   & \textbf{0.015} & \textbf{0.031} \\ \hline
\multirow{3}{*}{08} & ORB-SLAM2 & 11.46          & \textbf{0.28} & 43.09           & 0.192          & 0.061          \\
                    & DPT-VO    & 2.07           & 0.51          & 8.96            & 0.040          & 0.042          \\
                    & DF-VO     & \textbf{1.66}  & 0.33          & \textbf{6.96}   & \textbf{0.030} & \textbf{0.036} \\ \hline
\multirow{3}{*}{09} & ORB-SLAM2 & 9.30           & 0.26          & 38.77           & 0.128          & 0.061          \\
                    & DPT-VO    & \textbf{1.91}  & 0.39          & \textbf{6.93}   & 0.053          & 0.044          \\
                    & DF-VO     & 2.07           & \textbf{0.23} & 7.59            & \textbf{0.044} & \textbf{0.037} \\ \hline
\multirow{3}{*}{10} & ORB-SLAM2 & 2.57           & \textbf{0.32} & 5.42            & 0.045          & 0.065          \\
                    & DPT-VO    & \textbf{1.22}  & 0.42          & \textbf{1.55}   & \textbf{0.025} & 0.047          \\
                    & DF-VO     & 2.06           & 0.36          & 4.21            & 0.040          & \textbf{0.043} \\ \hline
\end{tabular}
\end{table}

Comparing DPT-VO with ORB-SLAM2 and DF-VO, according to Table~\ref{tab:metrics}, the DPT-VO achieved the best $t_{err}$ and ATE on sequences 00, 01, 05, 06, 09, and 10, with ATE also better for sequence 02. In addition, the $r_{err}$ obtained by DPT-VO was better than the other methods on sequences 00 and 06. As for rotation and translation RPEs, the DPT-VO model presented results very close to DF-VO, being better only for the translation RPE in sequence 10.

\begin{figure*}[tbp]
\centering
    \begin{subfigure}[]{0.24\textwidth}
        \includegraphics[width=0.97\textwidth]{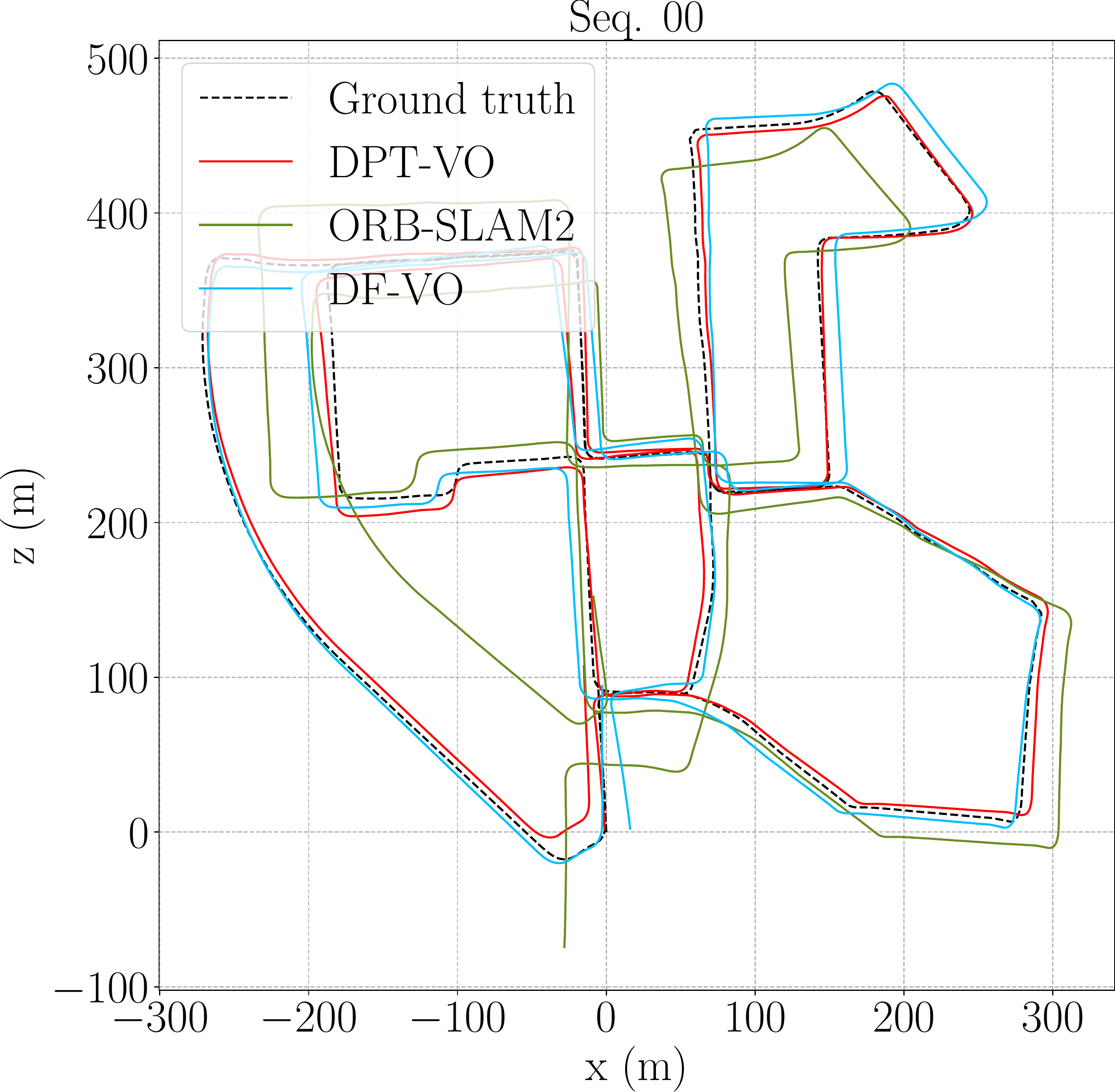}
    \end{subfigure}
    \hfill
    \begin{subfigure}[]{0.245\textwidth}
         \includegraphics[width=0.97\textwidth]{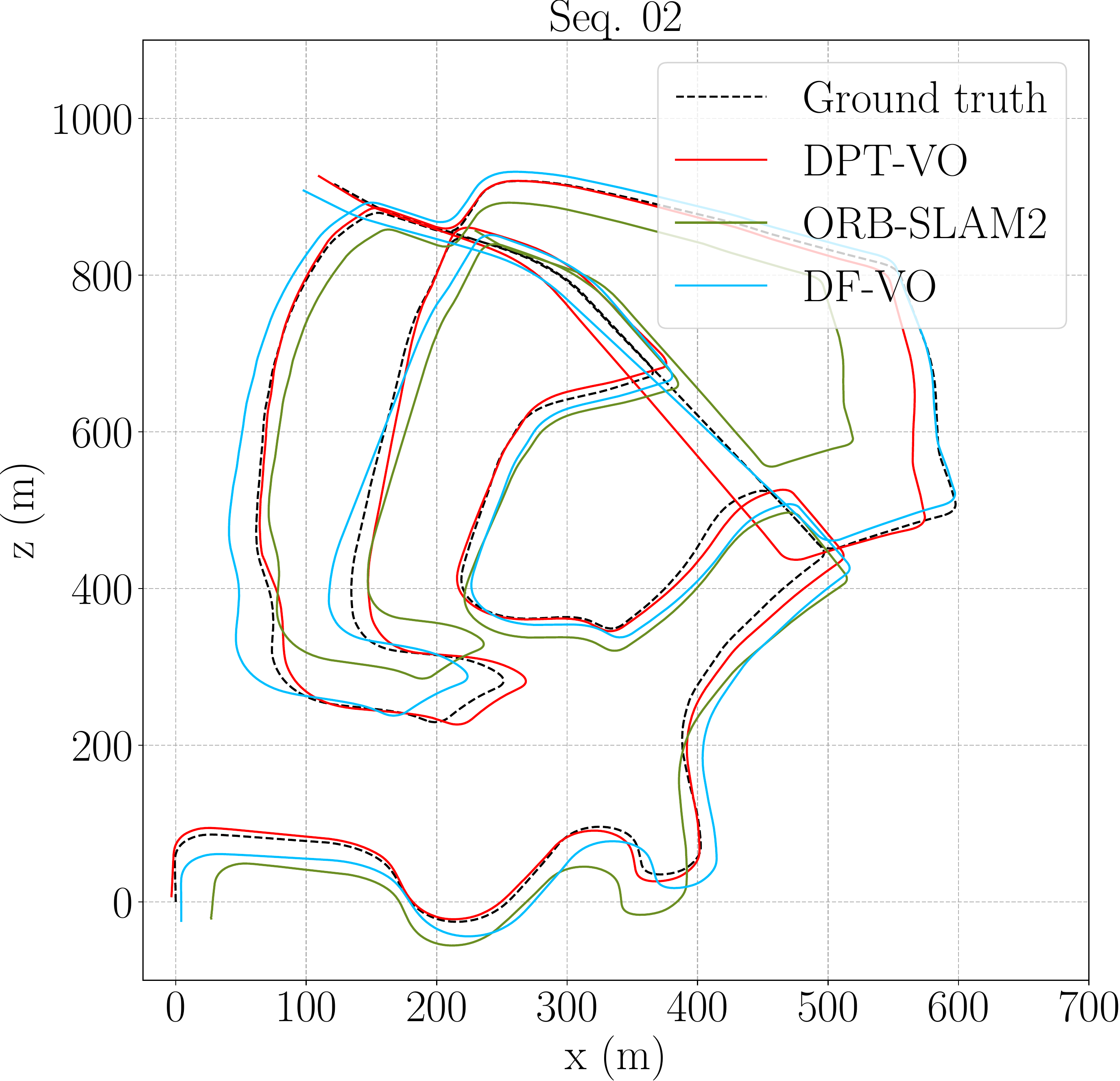}
    \end{subfigure}
    \hfill
    \begin{subfigure}[]{0.23\textwidth}
         \includegraphics[width=0.97\textwidth]{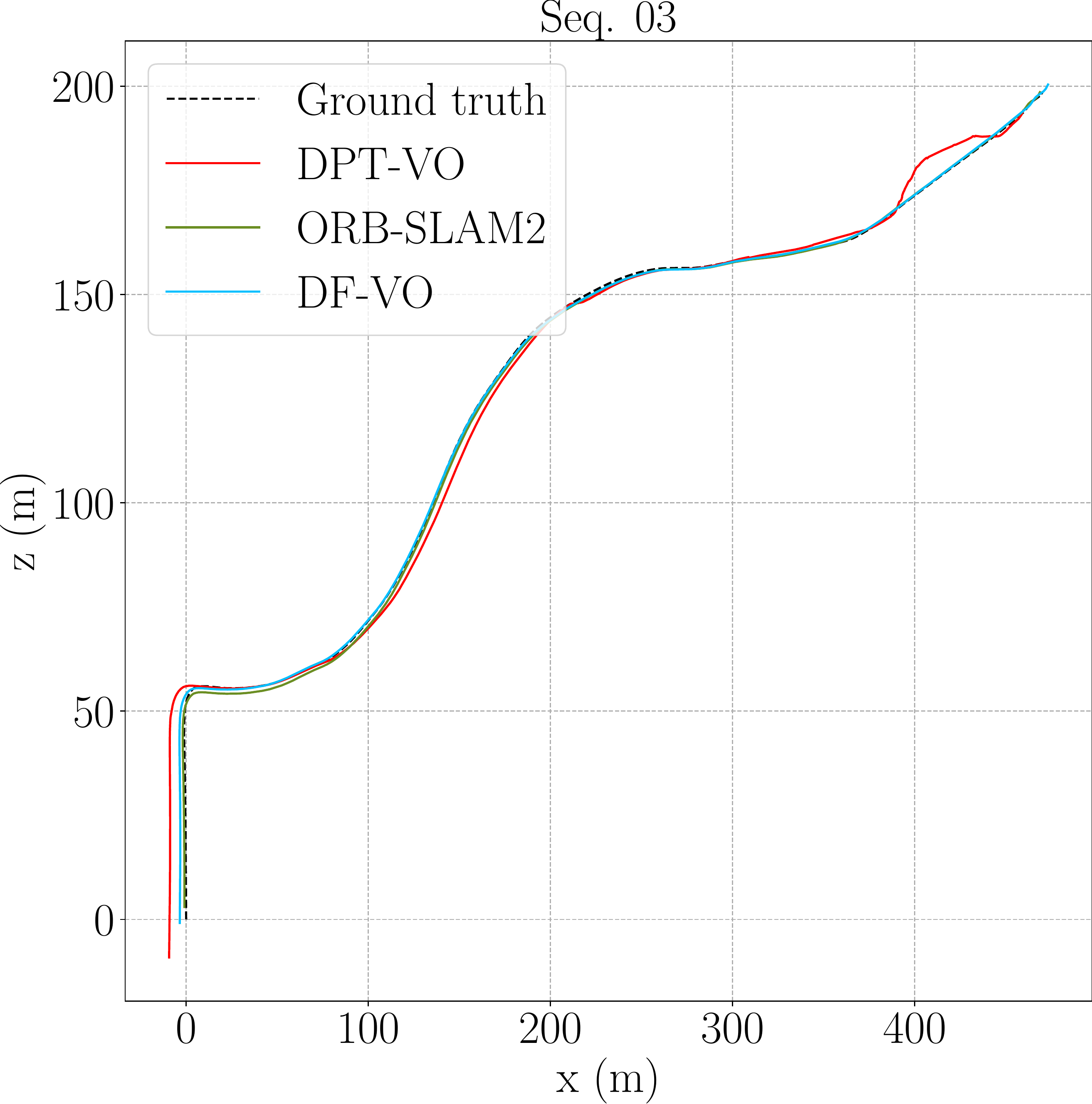}
    \end{subfigure}
    \hfill
    \begin{subfigure}[]{0.235\textwidth}
         \includegraphics[width=0.97\textwidth]{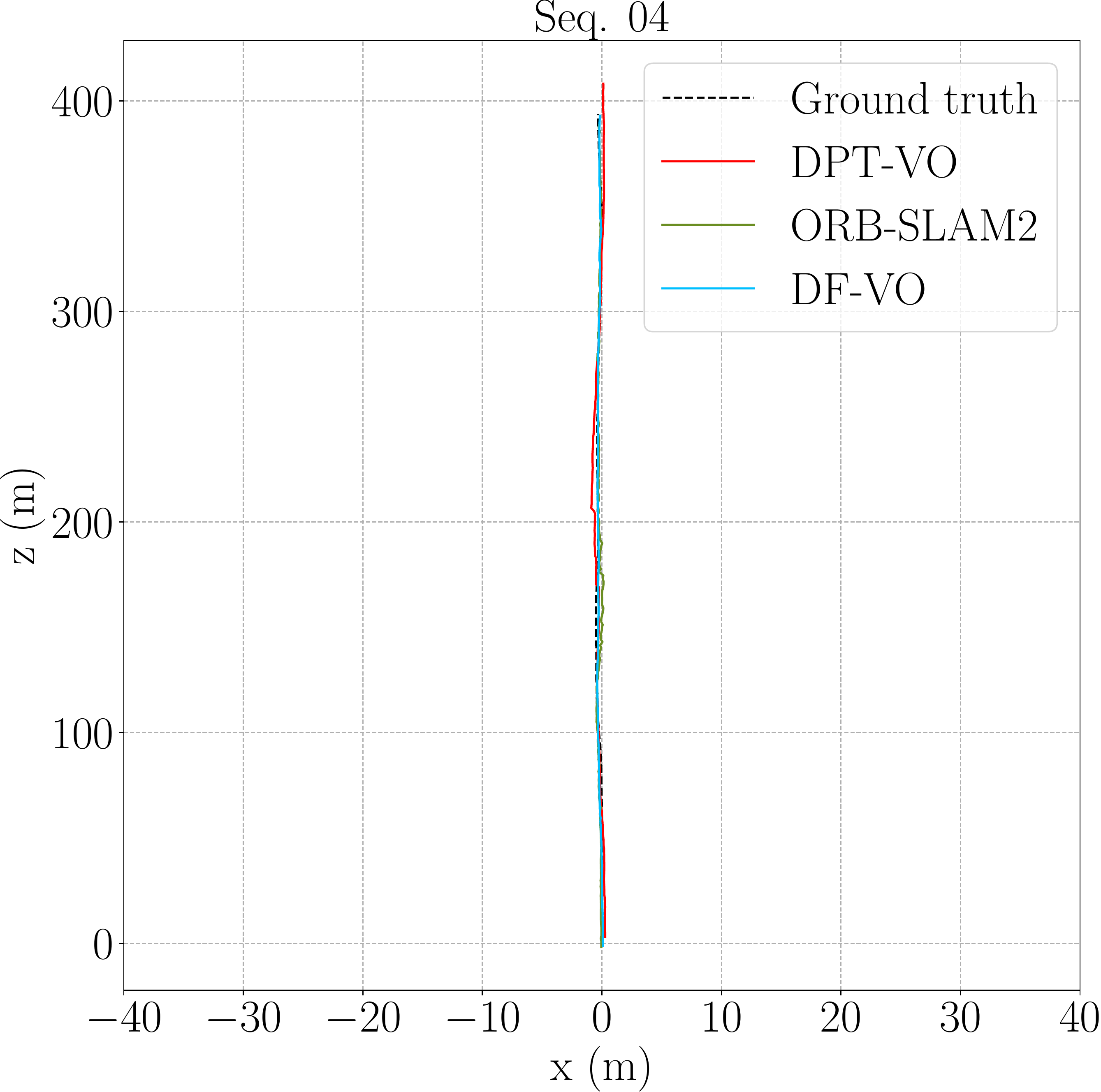}
    \end{subfigure}
    \par\bigskip
    \begin{subfigure}[]{0.24\textwidth}
         \includegraphics[width=0.97\textwidth]{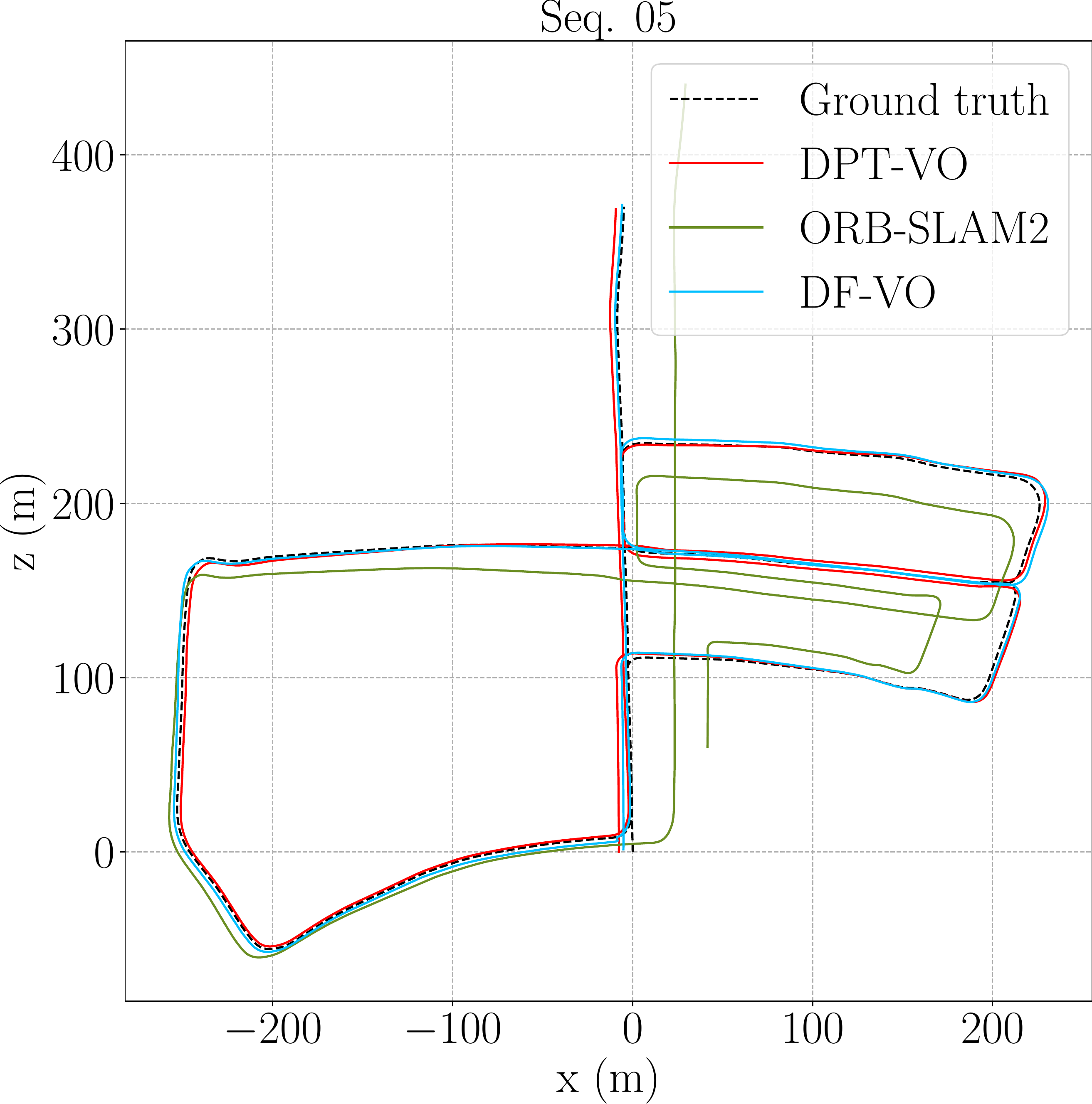}
    \end{subfigure}
    \hfill
    \begin{subfigure}[]{0.244\textwidth}
         \includegraphics[width=0.97\textwidth]{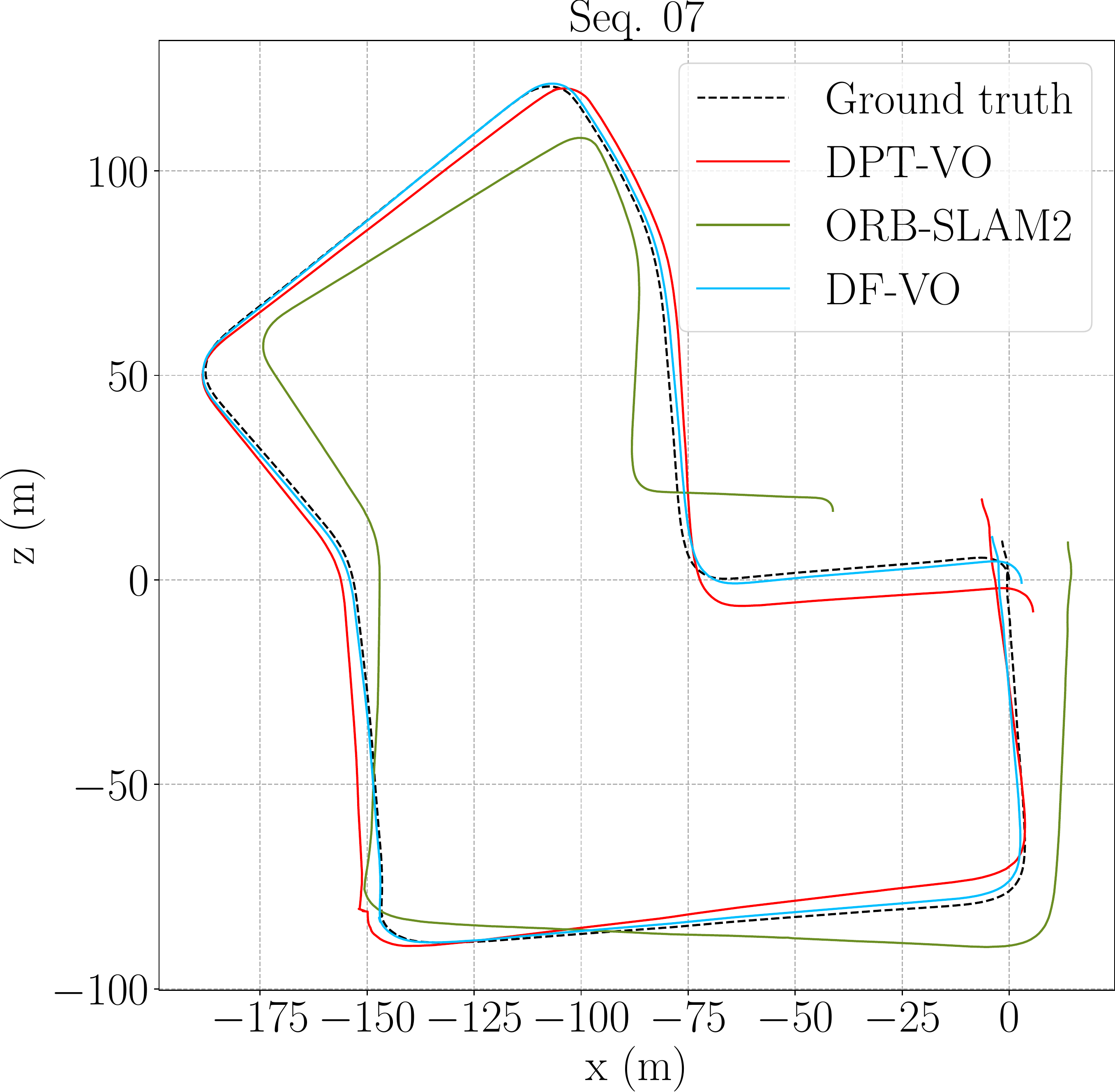}
    \end{subfigure}
    \hfill
    \begin{subfigure}[]{0.235\textwidth}
         \includegraphics[width=0.97\textwidth]{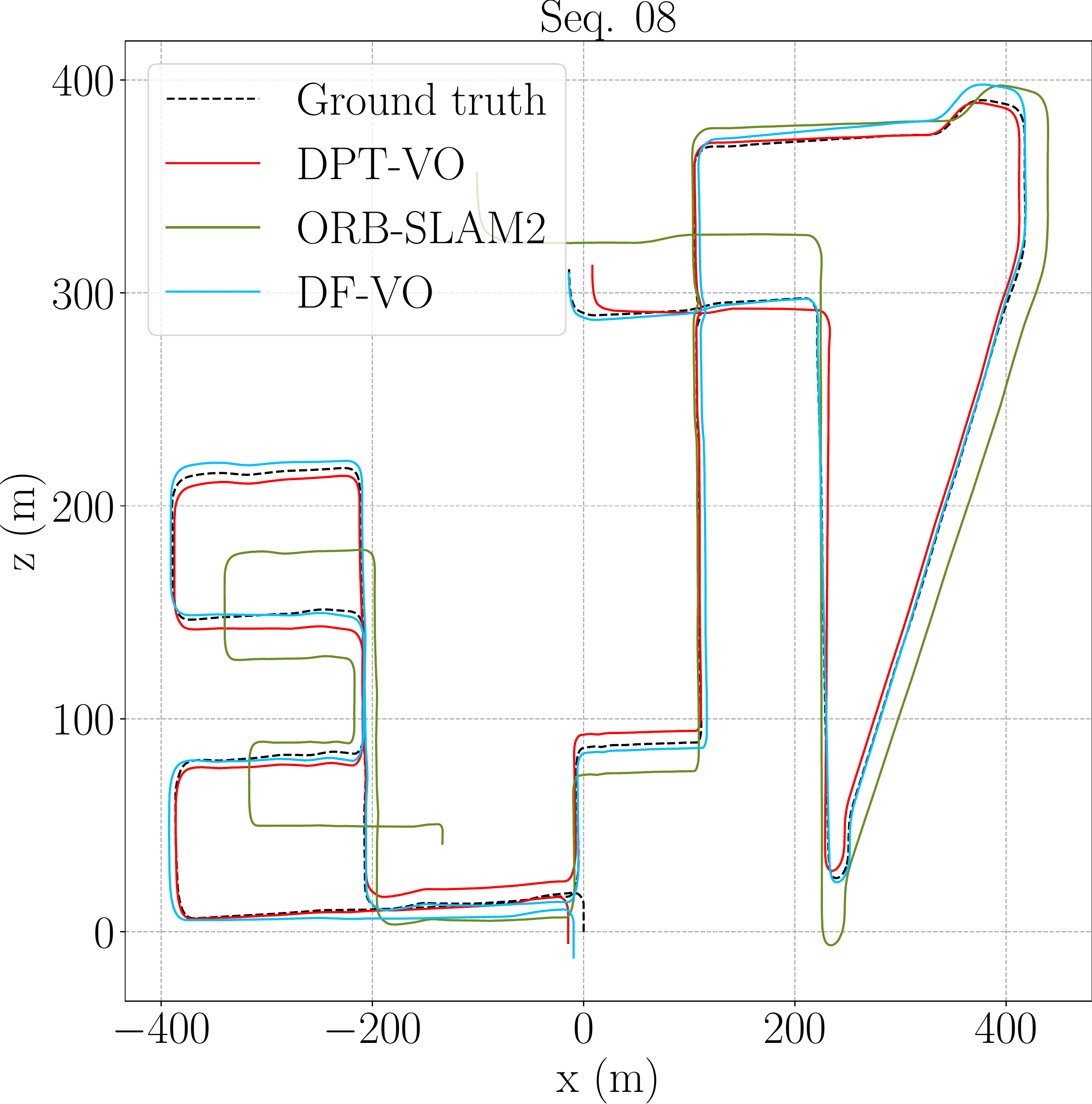}
    \end{subfigure}
    \hfill
    \begin{subfigure}[]{0.24\textwidth}
         \includegraphics[width=0.97\textwidth]{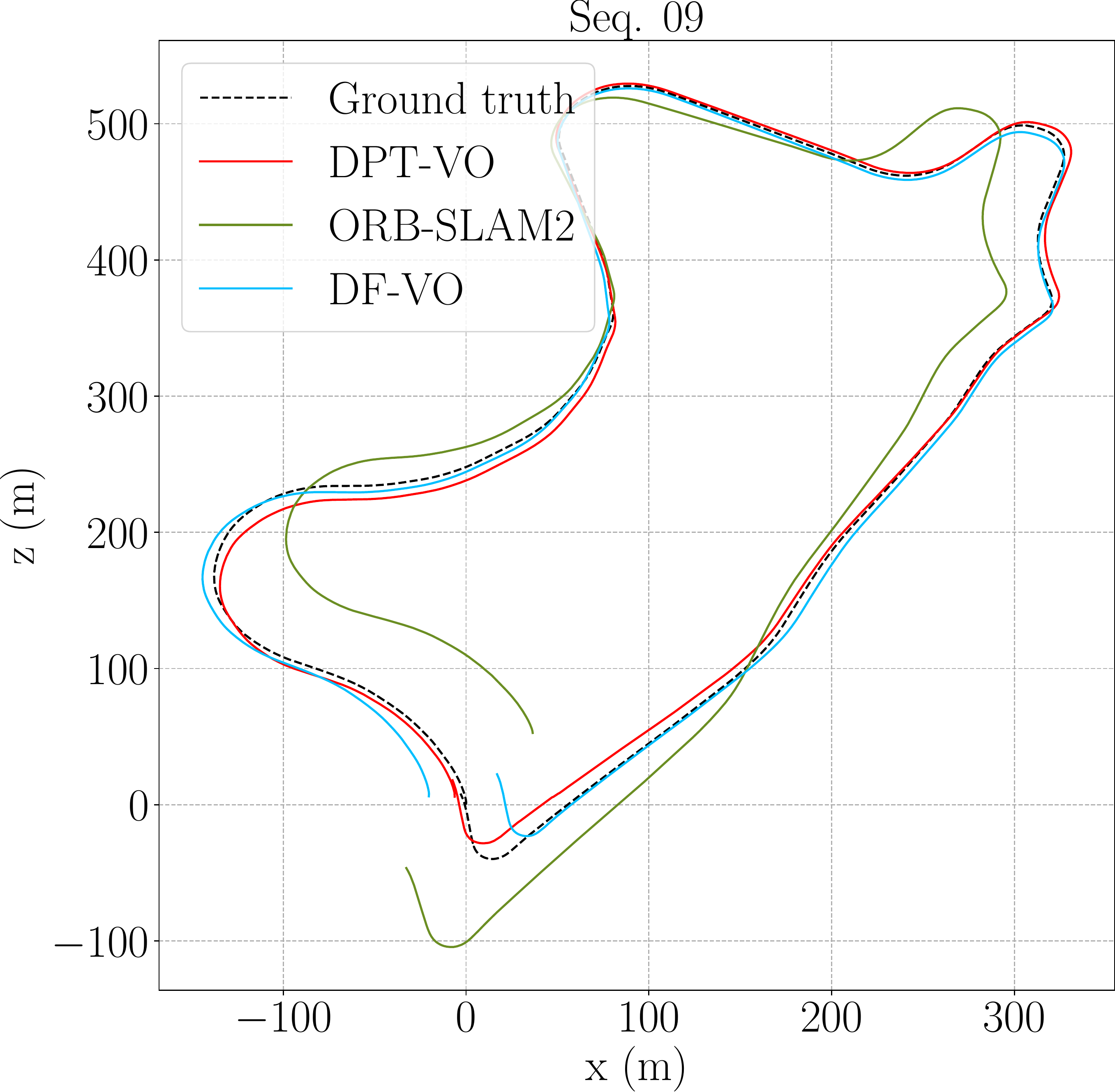}
    \end{subfigure}
    \caption{Qualitative results obtained by the ORB-SLAM2 without LC (\colorindicator{tab:olivedrab}), DF-VO (\colorindicator{tab:deepskyblue}), and DPT-VO (\colorindicator{tab:r}), compared with the ground truth (\colorindicator{tab:k}) in sequences 00, 02, 03, 04, 05, 07, 08, and 09 of the KITTI odometry dataset.}
    \label{fig:traj_comp}
\end{figure*}

Finally, to conclude our experiments, a further qualitative analysis was performed, now comparing the trajectories of the DPT-VO with the ORB-SLAM2 and DF-VO on eight sequences from the KITTI dataset, selected at random. Figure~\ref{fig:traj_comp} shows the trajectories obtained by the three algorithms on sequences 00, 02, 03, 04, 05, 07, 08, and 09 of the KITTI odometry dataset.
The qualitative results in Figure~\ref{fig:traj_comp} show that the trajectories obtained by DPT-VO were very close to the ground truths, likewise to the DF-VO trajectories. Also notice that the scale drift problem was significantly reduced in all the presented sequences. 

\section{Discussion}
\label{sec:discussion}

The presented results show that the transformer-based DPT model predicts the depth map from a single image, which can be used to reduce the scale ambiguity problem in MVO tasks. As shown in Figure~\ref{fig:plot_scale}, the scale drift problem was drastically reduced after the scale estimation step.

The quantitative analysis in Table 1 showed that DPT-VO was able to obtain the best ATE in 7 out of 11 KITTI sequences (63.6\%) and the best $t_{err}$ in 6 out of 11 sequences (54.5\%).
This result indicates that the translation errors were relatively small, mainly due to the scale drift reduction from the scale estimation step. The ORB-SLAM2 algorithm shows low rotation error ($r_{err}$) but high translation error ($t_{err}$), mainly due to the scale drift problem since we did not employ the loop closure. In general, the rotation errors of the DPT-VO were lower than the translation error, similar to the DF-VO results. Notice that sequence 01 has the worst metrics for all 3 methods, mainly the absolute trajectory error. This is caused by the high-speed scenario in this sequence, making it challenging to detect and track the features along the frames. 

The DF-VO method utilizes deep learning CNN-based models to predict the optical flow during the matching step and estimate the depth in monocular images, achieving high performance in MVO tasks. However, our DPT-VO model is simpler since it uses the classical Lucas-Kanade method to calculate the optical flow. Still, we achieved competitive results only by using a more accurate transformer-based model to predict the depth of monocular images, namely DPT. This means that the depth map estimated by a high precision and accurate model is essential to reduce the scale drift problems for the cases when depth map relations are used to estimate the scale in monocular systems. In other words, we showed that a transformer-based method has high potential to be used as a component of a MVO system, achieving competitive or even better results compared to CNN-based approaches. Our source code will be publicly available\footnote{https://github.com/aofrancani/DPT-VO}.

\section{Conclusion}
\label{sec:conclusion}

We presented an application of the dense prediction transformer for scale estimation in monocular visual odometry systems. The high performance of the DPT model in estimating the depth map from a single image contributed to an accurate scale estimation, reducing the scale drift in MVO tasks. Finally, the experimental results showed that our DPT-VO achieved competitive results compared to state-of-the-art methods on the KITTI odometry benchmark.

For future research, we plan to apply a transformer-based network for optical flow estimation, mainly in cases where traditional methods are not robust in detecting features due to lightning conditions. Moreover, we intend to propose an end-to-end network based on geometry and transformer approaches to estimate the camera pose in monocular visual odometry  problems.


\bibliographystyle{IEEEtran}
\bibliography{IEEEabrv,refs}

\end{document}